\pdfoutput=1

\documentclass[11pt]{article}

\usepackage[]{emnlp2021}

\usepackage{times}
\usepackage{latexsym}
\usepackage{url}
\usepackage[T1]{fontenc}
\usepackage{verbatim}
\usepackage{caption}
\usepackage{subfig}
\usepackage{svg}
\usepackage{graphicx}
\usepackage{tabularx}
\usepackage{balance}
\usepackage{hyperref}
\usepackage{multirow}
\usepackage{makecell}

\usepackage{algorithm}
\usepackage{latexsym}
\usepackage{algpseudocode}
\makeatletter
\usepackage{ amssymb }
\usepackage{amsmath}

\usepackage{booktabs}
\usepackage{cleveref}
\usepackage{enumitem}
\usepackage[T1]{fontenc}

\usepackage[utf8]{inputenc}

\usepackage{microtype}

\title{Back-Training excels Self-Training at Unsupervised Domain Adaptation of Question Generation and Passage Retrieval}

\author{
 Devang Kulshreshtha$^{1,2}$, Robert Belfer$^2$, Iulian Vlad Serban$^2$, Siva Reddy$^{1,3}$ \\
 $^1$Mila/McGill University\hspace{2em} $^2$Korbit Technologies \\
 $^3$Facebook CIFAR AI Chair\\
  {\small{\{devang.kulshreshtha, siva.reddy\}@mila.quebec}, \{iulian, robert\}@korbit.ai} \\
}
  
  \usepackage{todonotes}
\makeatletter
\newcommand*\iftodonotes{\if@todonotes@disabled\expandafter\@secondoftwo\else\expandafter\@firstoftwo\fi}  
\makeatother

\begin{document}
\maketitle

\begin{abstract}
In this work, we introduce back-training, an alternative to self-training for unsupervised domain adaptation (UDA) from source to target domain. While self-training generates synthetic training data where natural inputs are aligned with noisy outputs, back-training results in natural outputs aligned with noisy inputs. This significantly reduces the gap between the target domain and synthetic data distribution, and reduces model overfitting to the source domain. We run UDA experiments on question generation and passage retrieval from the \textit{Natural Questions} domain to machine learning and biomedical domains. We find that back-training vastly outperforms self-training by a mean improvement of 7.8 BLEU-4 points on generation, and 17.6\% top-20 retrieval accuracy across both domains. We further propose consistency filters to remove low-quality synthetic data before training. We also release a new domain-adaptation dataset- \textit{MLQuestions} containing 35K unaligned questions, 50K unaligned passages, and 3K aligned question-passage pairs.
\end{abstract}

\section{Introduction}


In domains such as education and medicine, collecting labeled data for tasks like question answering and generation requires domain experts, thereby making it expensive to build supervised models.
Transfer learning can circumvent this limitation by exploiting models trained on other domains where labeled data is readily available \cite{bengio2012deep,ruder2019transfer}.
However, using these pre-trained models directly without adapting to the target domain often leads to poor generalization due to distributional shift \cite{zhao2019learning}.
To address this issue, these models are further trained on cheap synthetically generated labeled data by exploiting unlabeled data from target domain \cite{ramponi-plank-2020-neural}.
One such popular data augmentation method for unsupervised domain adaptation (UDA) is \textit{self-training} \cite{yarowsky-1995-unsupervised}.

\begin{table}[t]
    \centering
    \begin{tabular}{lcc}
        \toprule
        
         &\multicolumn{2}{c}{\bf Synthetic Training Data} \\
        \bf Algorithm & Input & Output \\
        \midrule
        \multicolumn{3}{c}{\bf Question Generation (QG)} \\
        \midrule
     Self-Training & $p_u \sim P_{\mathcal{T}}(p)$ & $\hat{q} \sim P_{\mathcal{S}}(q|p_u)$  \\ 
     Back-Training & $\hat{p} \sim P_{\mathcal{S}}(p|q_u)$ & $q_u \sim P_{\mathcal{T}}(q)$  \\ 
        \midrule
        \multicolumn{3}{c}{\bf Passage Retrieval (IR)} \\
        \midrule
    Self-Training &  $q_u \sim P_{\mathcal{T}}(q)$ & $\hat{p} \sim P_{\mathcal{S}}(p|q_u)$  \\ 
    Back-Training &  $\hat{q} \sim P_{\mathcal{S}}(q|p_u)$ & $p_u \sim P_{\mathcal{T}}(p)$ \\ 
         \bottomrule
    \end{tabular}
    \caption{\textit{Self-Training} and \textit{Back-Training} for unsupervised domain adaptation of question generation and passage retrieval. In self-training, inputs are sampled from the target domain data distribution $P_{\mathcal{T}}$ and their corresponding outputs are generated using a supervised model $P_{\mathcal{S}}$ trained on the source domain. In back-training, the inverse happens: outputs are sampled from $P_{\mathcal{T}}$ and their corresponding inputs are generated using $P_{\mathcal{S}}$. Notation: $q$ and $p$ denote questions and passages respectively, $._{u}$ denotes samples from the target domain and $\hat{.}$ denotes the samples generated by a supervised model trained on the source domain.}
    \label{tab:STvsCT-main}
\end{table}
In self-training, given a pre-trained model that can perform the task of interest in a source domain and unlabeled data from the target domain, the pre-trained model is used to predict noisy labels for the target domain data.
The pre-trained model is then fine-tuned on synthetic data to adapt to the new domain.
To improve the quality of the synthetic data, it is also common to filter out low-confidence model predictions \cite{zhu2005semi}.

A model fine-tuned on its own confidence predictions might suffer from confirmation bias which leads to overfitting \cite{yu2020ensemble}. This means that the distributional gap between the target domain's true output distribution and the learned output distribution could grow wider as training proceeds. 
In this paper, we propose a new training protocol called \textit{back-training} which closes this gap (the name is inspired from \textit{back-translation} for machine translation).
While self-training generates synthetic data where noisy outputs are aligned with quality inputs, back-training generates quality outputs aligned with noisy inputs. The model fine-tuned to predict real target domain outputs from noisy inputs 
reduces overfitting to the source domain \cite{vincent2008extracting}, and matches the target domain distribution more closely.

We focus on unsupervised domain adaptation (UDA) of Question Generation (QG) and Passage Retrieval (IR) from generic domains such as Wikipedia to target domains.
Our target domain of interest is \textit{machine learning}, as it is a rapidly evolving area of research. QG and IR tasks could empower student learning on MOOCs \cite{heilman2010good}.
For example, from a passage about linear and logistic regression, an education bot could generate questions such as \textit{what is the difference between linear and logistic regression?} to teach a student about these concepts.
Moreover, IR models could help students find relevant passages for a given question \cite{fernandez2009teaching}.
In this domain, unsupervised data such as text passages and questions are easy to obtain separately rather than aligned to each other.

We also perform our main domain adaptation experiments on \textit{biomedical} domain using PubMedQA dataset \cite{jin2019pubmedqa} to further strengthen our hypothesis.

    
\Cref{tab:STvsCT-main} demonstrates the differences between self-training and back-training for QG and IR.
Consider the QG task:
for self-training, we first train a supervised model $P_{\mathcal{S}}(q|p)$ on the source domain that can generate a question $q$ given a passage $p$.
We use this model to generate a question $\hat{q}$ for an unsupervised passages $p_u$ sampled from the target domain distribution $P_{\mathcal{T}}(p)$.
Note that $\hat{q}$ is generated conditioned on the target domain passage using $P_\mathcal{S}(q|p_u)$.
We use the pairs ($p_u$, $\hat{q}$) as the synthetic training data to adapt $P_S(q|p)$ to the target domain.
In back-training, we assume access to unsupervised questions and passages from the target domain.
We first train an IR model $P_S(p|q)$ on the source domain, then sample a question $q_u$ from the target domain distribution $P_\mathcal{T}(q)$.
We condition the retriever on this question  i.e., $P_S(p|q_u)$, and retrieve a passage $\hat{p}$ from the target domain and treat it as a noisy alignment.
We use the pairs ($\hat{p}$, $q_u$) as the synthetic training data to adapt $P_S(q|p)$.
\Cref{tab:STvsCT-main} also describes the details of domain adaptation for the passage retriever.

Our contributions and findings are as follows:
1)~We show that QG and IR models trained on NaturalQuestions \cite{kwiatkowski2019natural} generalize poorly to target domains, with at least 17\% mean performance decline on both QG and IR tasks.
2)~Although self-training improves the domain performance marginally, our back-training method outperforms self-training by a mean improvement of 7.8 BLEU-4 points on generation, and 17.6\% top-20 retrieval accuracy across both target domains.
3) We further propose consistency filters to remove low-quality synthetic data before training.
4)~We release \textit{MLQuestions}: a domain adaptation dataset for the machine learning domain containing 35K unaligned questions, 50K unaligned passages, and 3K aligned question-passage pairs.





\section{Background}
\label{sec:background}

In this section, we describe the source and target domain datasets, models for question generation and passage retrieval, and the evaluation metrics.

\begin{table*}[]
    \centering
    \footnotesize
    \begin{tabularx}{\textwidth}{lcccc}
        \Xhline{2\arrayrulewidth}
        \multirow{2}{*}{\makecell{\textbf{Taxonomy}}} & \multirow{2}{*}{\makecell{\textbf{Examples}\\\textbf{(from $\textbf{MLQuestions}$)}}} & \multirow{2}{*}{\makecell{\textbf{Description}\\\textbf{\textit{(Frequent Wh-words)}}}} & \multicolumn{2}{c}{\textbf{Distribution (\%)}} \\
        & & & \textbf{NaturalQuestions} & \textbf{MLQuestions}\\
        \hline
        \multirow{3}{*}{\makecell{DESCRIPTION}} & \multirow{3}{*}{\makecell{ \underline{What} is supervised \\learning with example?}} & \multirow{3}{*}{\makecell{Asking definition or \\examples about a concept\\\textit{(What, Who, When, Where)}}} & \multirow{3}{*}{86\%} & \multirow{3}{*}{39\%}\\
         & & & \\
         & & & \\
         \hline
         
         \multirow{2}{*}{\makecell{METHOD}} & \multirow{2}{*}{\makecell{\underline{How} do you compute \\vectors in Word2Vec?}} & \multirow{2}{*}{\makecell{Computational or procedural \\questions - \textit{(How)}}} & \multirow{2}{*}{1\%} & \multirow{2}{*}{15\%}\\
         & & & \\
         \hline
         
         \multirow{2}{*}{\makecell{EXPLANATION}} & \multirow{2}{*}{\makecell{\underline{Why} does ReLU activation \\work so surprisingly well?}} & \multirow{2}{*}{\makecell{Causal, justification or\\ goal-oriented questions - \textit{(Why)}}} & \multirow{2}{*}{3\%} & \multirow{2}{*}{18\%}\\
          & & & \\
          \hline
          
           \multirow{2}{*}{\makecell{COMPARISON}} & \multirow{2}{*}{\makecell{\underline{What is the difference} \\\underline{between} LDA and PCA?}} & \multirow{2}{*}{\makecell{Ask to compare more than\\ one concept with each other}} & \multirow{2}{*}{5\%} & \multirow{2}{*}{10\%}\\
          & & & \\
          \hline
          
          \multirow{2}{*}{\makecell{PREFERENCE}} & \multirow{2}{*}{\makecell{\underline{Is} language acquisition \\innate or learned?}} & \multirow{2}{*}{\makecell{Yes/No or select from valid\\ set of options - \textit{(Is, Are)}}} & \multirow{2}{*}{5\%} & \multirow{2}{*}{18\%}\\
          & & & \\
        \Xhline{2\arrayrulewidth}
    \end{tabularx}
    \caption{Classification of 200 random questions from NaturalQuestions and MLQuestions as per \citet{nielsen2008question}.}
    \label{tab:taxonomy-table}
\end{table*}

\subsection{Source Domain: NaturalQuestions}\label{sec:nq}
We use the NaturalQuestions dataset \cite{kwiatkowski2019natural} as our source domain.
NaturalQuestions is an open-domain question answering dataset containing questions from Google search engine queries paired with answers from Wikipedia.
We use the long form of the answer which corresponds to passages (paragraphs) of Wikipedia articles.
It is the largest dataset available for open-domain QA, comprising of 300K training examples, each example comprising of a question paired with a Wikipedia passage.
We label 200 random questions of NaturalQuestions and annotate them into 5 different classes based on the nature of the question as per \citet{nielsen2008taxonomy}. \Cref{tab:taxonomy-table} shows these classes and their distribution.
As seen, 86\% of them are descriptive questions starting with \textit{what, who, when} and \textit{where}. Refer to \Cref{sec:nq-preprocess} for details on dataset pre-processing and \Cref{sec:taxonomy} for detailed taxonomy description.

\subsection{Target Domain I: Machine Learning}\label{sec:mlquestions}
Our first target domain of interest is machine learning. There is no large supervised QA dataset for this domain, and it is expensive to create one since it requires domain experts. However, it is relatively cheap to collect a large number of ML articles and questions. We collect ML concepts and passages from the Wikipedia machine learning page\footnote{\url{https://en.wikipedia.org/wiki/Category:Machine_learning}} and recursively traverse its subcategories. We end up with 1.7K concepts such as \textit{Autoencoder},  \textit{word2vec} etc. and 50K passages related to these concepts.

For question mining, we piggy-back on Google Suggest's \textit{People also ask} feature to collect 104K questions by using above machine learning concept terms as seed queries combined with question terms such as \textit{what}, \textit{why} and \textit{how}.
However, many questions could belong to generic domain due to ambiguous terms such as \textit{eager learning}.
We employ three domain experts to annotate 1000 questions to classify if a question is in-domain or out-of-domain. 
Using this data, we train a classifier \cite{liu2019roberta} to filter questions that have in-domain probability less than 0.8.
This resulted in 46K in-domain questions, and has 92\% accuracy upon analysing 100 questions.
Of these, we use 35K questions as unsupervised data. See \cref{ood_clf} for classifier training details and performance validation.

The rest of the 11K questions are used to create supervised data for model evaluation.
We use the Google search engine to find answer passages to these questions, resulting around 11K passages.
Among these, we select 3K question and passage pairs as the evaluation set for QG (50\% validation and 50\% test).
For IR, we use the full 11K passages as candidate passages for the 3K questions.
We call our dataset \textit{MLQuestions}.

\Cref{tab:taxonomy-table} compares MLQuestions with NaturalQuestions.
\hspace{-0.1em}We note that MLQuestions has higher diversity of question classes than NaturalQuestions, making the transfer setting challenging.
\subsection{Target Domain II: Biomedical Science}\label{sec:pubmedqa}
Our second domain of interest is biomedicine for which we use PubMedQA \cite{jin2019pubmedqa} dataset. Questions are extracted from PubMed abstract titles ending with question mark, and passages are the conclusive part of the abstract. As unsupervised data, we utilize PQA-U(nlabeled) subset containing 61.2K unaligned questions and passages. For supervised data, we use PQA-L(abeled) subset of 1K question-passage pairs manually curated by domain experts. We use the same dev-test split of 50-50\% as \cite{jin2019pubmedqa} as the evaluation set for QG. 
For IR, in order to have the same number of candidate passages as MLQuestions, we combine randomly sampled 10K passages from PQA-U with 1K PQA-L passages to get 11K passages as candidate passages for 1K questions.

\subsection{Question Generation Model}
We use BART \cite{lewis2020bart} to train a supervised QG model on NaturalQuestions.
BART is a Transformer encoder-decoder model pretrained to reconstruct original text inputs from noisy text inputs.
Essentially for QG, BART is further trained to learn a conditional language model $P_\mathcal{S}(q|p)$ that generates a question $q$ given a passage $p$ from the source domain. For experimental details, see \ref{sec:train-details}.

\subsection{Passage Retrieval Model}
We use the pretrained Dense Passage Retriever (DPR; \citealt{karpukhin2020dense}) on NaturalQuestions.
DPR encodes a question $q$ and passage $p$ separately using a BERT bi-encoder and is trained to maximize the dot product (similarity) between the encodings $E_P(p)$ and $E_Q(q)$, while minimizing similarity with other closely related but negative passages.
Essentially, DPR is a conditional classifier $P_S(p|q)$ that retrieves a relevant passage $p$
given a question $q$ from the source domain. For model training details, see \ref{sec:train-details}.

\subsection{Evaluation Metrics}\label{sec:eval-metrics}
We evaluate question generation using standard language generation metrics: BLEU1-4 \cite{papineni2002bleu}, METEOR \cite{banerjee2005meteor} and ROUGE$_{L}$ \cite{lin2004rouge}. They are abbreviated as B1, B2, B3, B4, M, and R respectively throughout the paper.
We also perform human evaluation on the model generated questions. 
For passage retrieval, we report top-k retrieval accuracy for $k = 1, 10, 20, 40, 100$ following \citet{karpukhin2020dense} by measuring the fraction of cases where the correct passage lies in the top $k$ retrieved passages.
We consider 11K passages in all datasets for retrieval during test time.

\section{Transfer from Source to Target Domain without Adaptation}
\begin{figure}
\vspace{.3cm}
    \includegraphics[trim=3.3cm 19.4cm 5cm 4.2cm, clip, scale=0.9]{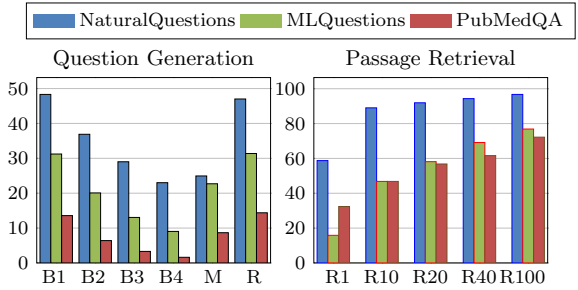}
    \caption{IID/OOD generalization gaps for Question Generation and Passage Retrieval due to distributional shift between source and target domains. For a fair comparison, the number of candidate passages for IR are kept similar for all datasets.}
    \label{fig:iid-ood-drop}
\end{figure}

We investigate how well models trained on NaturalQuestions transfer directly to our target domains without any domain adaptation.
For comparison, we also present the results on NaturalQuestions.
To be fair, we sample equal number of samples from the development set of NaturalQuestions as in the test set of MLQuestions and PubMedQA for QG and IR tasks.
Figure \ref{fig:iid-ood-drop} shows the results.
We observe high performance drops across all generation metrics (14-20\%) from NaturalQuestions (IID data) to MLQuestions and PubMedQA (OOD Data).
Human evaluation on QG (see Table \ref{tab:human_evalutation}) also reveals that the generated questions are either generic, or fail to understand domain-specific terminology. 
OOD performance in the IR task is even worse (25-40\% drop), revealing a huge distribution shift between the source and target domain.

\section{Unsupervised Domain Adaptation}
In this section, we describe self-training and back-training methods to generate synthetic training data for unsupervised domain adaptation (UDA).
We also introduce consistency filters to further improve the quality of the synthetic data.

\subsection{Problem Setup}
The source domain consists of labeled data containing questions paired with passages $\mathcal{D}_{\mathcal{S}} \equiv \{(q_s^i,p_s^i)\}_{i=1}^m$.
The target domain consists of unlabeled passages $\mathcal{P}_{\mathcal{U}} \equiv \{p_u^i\}_{i=1}^{m_p}$, and unlabeled questions $\mathcal{Q}_{\mathcal{U}} \equiv \{q_u^i\}_{i=1}^{m_q}$. 
Note that $\mathcal{P}_{\mathcal{U}}$ and $\mathcal{Q}_{\mathcal{U}}$ are \textit{not necessarily} aligned with each other. 
Given this setup, our goal is to learn QG and IR models with parameters $\boldsymbol{\theta} \equiv \{\boldsymbol{\theta}_G, \boldsymbol{\theta}_R\}$ that can achieve high generation and retrieval performance on target domain~$T$. 
Table \ref{tab:notation} describes the notations used across the paper.

\begin{table}[]
    \centering
    \small
    \begin{tabular}{ll}
        \hline
        \textbf{Notation} & \textbf{Definition} \\
        \hline
        $S,T$ & Source, Target Domain \\
        $P_S, P_T$ & Source, Target data distribution \\
        $\mathcal{D}_{\mathcal{S}} \equiv \{(q_s^i,p_s^i)\}_{i=1}^m$ & Source labeled corpus \\
        $\mathcal{P}_{\mathcal{U}} \equiv \{p_u^i\}_{i=1}^{m_p}$ & Target unlabeled passages \\
        $\mathcal{Q}_{\mathcal{U}} \equiv \{q_u^i\}_{i=1}^{m_q}$ & Target unlabeled questions \\
        $\boldsymbol{\theta} \equiv \{\boldsymbol{\theta}_G, \boldsymbol{\theta}_R\}$ & QG, IR Models \\
        $S_G, S_R$ & Synthetic data for QG, IR \\
        \hline
    \end{tabular}
    \caption{Notations used throughout the paper.}
    \label{tab:notation}
\end{table}

\subsection{Self-Training for UDA}\label{self-training}
Self-training \cite{yarowsky-1995-unsupervised} involves training a model on its own predictions.
We present the proposed self-training for UDA in \Cref{algo}. 
First the baseline models $\boldsymbol{\theta_{G}}$ and $\boldsymbol{\theta_{R}}$ are trained on the source passage-question corpus $\mathcal{D}_{\mathcal{S}}$. 
Then, at each iteration, the above models generate pseudo-labeled data from unlabeled passages $\mathcal{P}_{\mathcal{U}}$ for question generation and questions $\mathcal{Q}_{\mathcal{U}}$ for passage retrieval.
For QG, $\boldsymbol{\theta_{G}}$ generates a question $\hat{q}$ for each $p_u \in \mathcal{P}_{\mathcal{U}}$ and adds $(p_u,\hat{q})$ to synthetic data $S_G$.
For IR, $\boldsymbol{\theta_{R}}$ retrieves a passage $\hat{p}$ from $\mathcal{P}_{\mathcal{U}}$ for each $q_u \in \mathcal{Q}_{\mathcal{U}}$ and adds $(q_u,\hat{p})$ to $S_R$. 
The models $\boldsymbol{\theta_{G}}$ and $\boldsymbol{\theta_{R}}$ are fine-tuned on $S_G$ and $S_R$ respectively. The process is repeated for a desired number of iterations, which we refer to as \textit{iterative refinement}.
Note that in self-training, inputs are sampled from target domain and the outputs are predicted (noisy).

\subsection{Back-Training for UDA}\label{cross-task-co-training}
The main idea of back-training is to work backwards: start with true output samples from the target domain, and predict corresponding inputs which aligns the most with the output.
While self-training assumes inputs are sampled from the target domain distribution, back-training assumes outputs are sampled from the target domain distribution.
When two tasks are of dual nature (i.e., input of one task becomes the output of another task), back-training can be used to generate synthetic training data of one task using the other, but on a condition that outputs can be sampled from the target domain distribution.
QG and IR tasks meet both criteria.
For QG, we have unlabeled questions in the target domain and its dual friend IR can retrieve their corresponding input passages from the target domain. For IR, we have passages  in the target domain and QG can generate their input questions.
Formally, for QG, the IR model $\boldsymbol{\theta_{R}}$ retrieves passage $\hat{p}$ from $\mathcal{P}_{\mathcal{U}}$ for each $q_u \in \mathcal{Q}_{\mathcal{U}}$ and adds $(\hat{p},q_u)$ to $S_G$.
For IR, the QG model $\boldsymbol{\theta_{G}}$ generates a question $\hat{q}$ for each $p_u \in \mathcal{P}_{\mathcal{U}}$ and adds $(\hat{q},p_u)$ to $S_R$. 

\paragraph{Similarities with \textit{back-translation}}
Back-translation is an effective method to improve machine translation using synthetic parallel corpora containing human-produced target language sentences paired with artificial source language translations \cite{sennrich-etal-2016-improving,edunov2018understanding}. 
Back-training is inspired by this idea, however it is not limited to machine translation. 

\begin{algorithm}[t]
\begin{algorithmic}[1]
\footnotesize
\Require {
Source Data $\mathcal{D}_{\mathcal{S}} \equiv \{(q_s^i, p_s^i)\}_{i=1}^{m}$, Target unlabeled data  $\mathcal{P}_{\mathcal{U}} \equiv \{p_u^i\}_{i=1}^{m^p}$, $\mathcal{Q}_{\mathcal{U}} \equiv \{q_u^i\}_{i=1}^{m_q}$
}
\Ensure Target domain QG model $\boldsymbol{\theta_{G}}$, IR model $\boldsymbol{\theta_{R}}$
\State \textbf{Init:} $\boldsymbol{\theta_{G}}, \boldsymbol{\theta_{R}} \leftarrow $ Train on $\mathcal{D}_{\mathcal{S}}$
\Repeat 
\State $S_G \leftarrow [\ ], S_R \leftarrow [\ ]$ \Comment{Synthetic data for $\boldsymbol{\theta_{G}}$ and $\boldsymbol{\theta_{R}}$}
\For{$q_u \in \mathcal{Q}_{\mathcal{U}}$}
\State $\hat{p} \leftarrow $ Retrieve $p$ from $\mathcal{P}_{\mathcal{U}}$ closest to $q_u$ using $\boldsymbol{\theta_{R}}$
\State add $(\hat{p}, q_u)$ to \colorbox{red!50}{$S_{R}$} \colorbox{blue!50}{$S_{G}$}
\EndFor
\For{$p_u \in \mathcal{P}_{\mathcal{U}}$}
\State $\hat{q} \leftarrow $ Generate $q$ from $p_u$ using $\boldsymbol{\theta_{G}}$
\State add $(p_u, \hat{q})$ to \colorbox{red!50}{$S_G$} \colorbox{blue!50}{$S_{R}$}
\EndFor
\State $ \boldsymbol{\theta_{G}} \leftarrow$ Finetune on $S_G$, $\boldsymbol{\theta_{R}} \leftarrow$ Finetune on $S_R$
\Until{\textit{dev performance decreases}}
\end{algorithmic}
\caption{\small{Vanilla \colorbox{red!50}{Self-Training} \colorbox{blue!50}{Back-Training} for unsupervised domain adaptation. Vanilla algorithms can be improved further using consistency~filters}}
\label{algo}
\end{algorithm}

\begin{figure}
    \centering
    \includegraphics[trim=0.075cm 23cm 10cm 0.515cm, clip, scale=0.825]{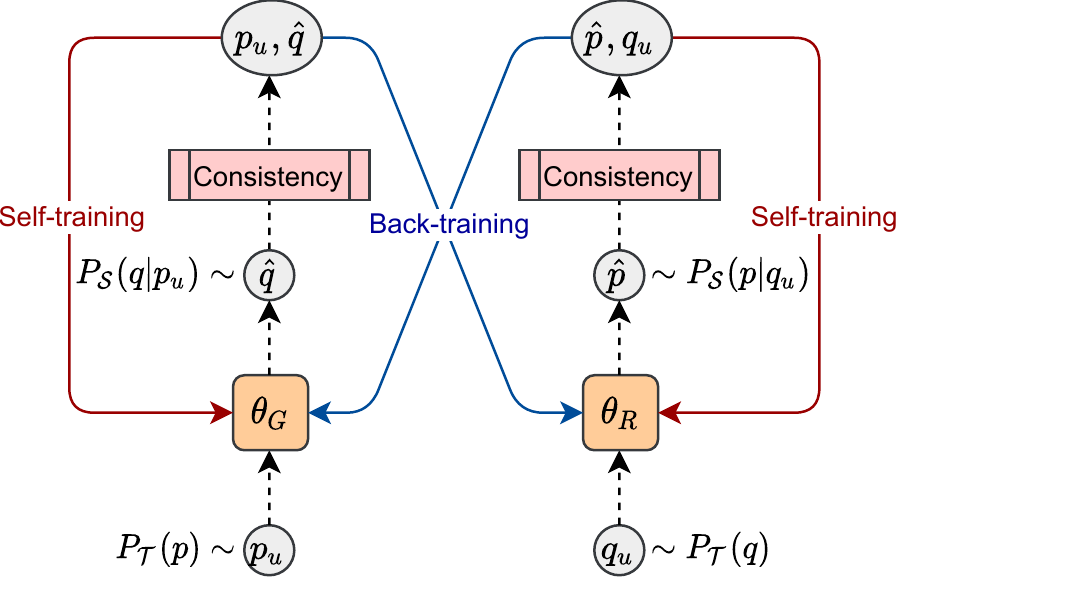}
    \caption{\textcolor{red}{Self-training} and \textcolor{blue}{Back-training} for UDA.}
    \label{fig:combinations}
\end{figure}

\begin{table*}[ht]
\setlength{\tabcolsep}{5.25pt}
\footnotesize
    \centering
    \begin{tabular}{c|c|cccccc|cccc}
    \toprule
        \multirow{2}{*}{\textbf{Dataset}} & \multirow{2}{*}{\textbf{Model}} & \multicolumn{6}{c|}{\textbf{Question Generation}} & \multicolumn{4}{c}{\textbf{Passage Retrieval}} \\
        \cline{3-12}
         & & B1 & B2 & B3 & B4 & M & R & R@1 & R@20 & R@40 & R@100 \\
        \hline
       \multirow{3}{*}{\textit{MLQuestions}} & No-adaptation & 31.23 & 20.07 & 13.05 & 9.04 & 22.70 & 31.38 & 15.86 & 58.13 & 69.13 & 76.86 \\
        & Self-Training & 31.81 & 20.74 & 13.61 & 9.43 & 23.31 & 32.18 & 17.86 & 65.26 & 74.13 & 83.06 \\
        & \textbf{Back-Training} & \textbf{44.12} & \textbf{32.86} & \textbf{24.21} & \textbf{18.48} & \textbf{23.83} & \textbf{43.97} & \textbf{24.53} & \textbf{77.73} & \textbf{84.8} & \textbf{91.73} \\
        \hline
        \multirow{3}{*}{\textit{PubMedQA}} & No-adaptation & 13.57 & 6.41 & 3.31 & 1.62 & 8.67 & 14.38 & 32.4 & 56.8 & 61.6 & 72.2 \\
        & Self-Training & 13.36 & 6.28 & 3.25 & 1.64 & 8.84 & 15.00 & 32.8 & 57.0 & 63.6 & 72.8 \\
        & \textbf{Back-Training} & \textbf{26.71} & \textbf{17.01} & \textbf{11.80} & \textbf{8.25} & \textbf{16.99} & \textbf{25.14} & \textbf{55.4} & \textbf{79.8} & \textbf{81.8} & \textbf{85.8}
        \\
        \bottomrule
        \hline
    \end{tabular}
    \caption{Results of unsupervised domain adaptation. \textit{No-adaptation} denotes the model trained on NaturalQuestions and tested directly on MLQuestions/PubMedQA without any domain adaptation.}
    \label{tab:aqgm}
\end{table*}

\subsection{Consistency filters for Self-Training and Back-Training}
\label{sec:filters}

The above algorithms utilize \textit{full} unlabeled data along with their predictions even if the predictions are of low confidence.
To alleviate this problem, in self-training, it is common to filter low-confidence predictions \cite{zhu2005semi}.
We generalize this notion as \textit{consistency filtering}:
For the tasks QG and IR, a \textit{generator} $G \in \{\boldsymbol{\theta}_G, \boldsymbol{\theta}_R\}$ produces synthetic training data for a task whereas the \textit{critic} $C \in \{\boldsymbol{\theta}_G, \boldsymbol{\theta}_R\}$ filters low confidence predictions. 
We define two types of consistency filtering: 1) \textbf{Self consistency} where the generator and critic are the \textit{same}. This is equivalent to filtering out model's own low confidence predictions in self-training. 2) \textbf{Cross consistency} where the generator and critic are \textit{different}. This means $\boldsymbol{\theta_R}$ will filter the synthetic data generated by $\boldsymbol{\theta_G}$, and vice-versa. For $\boldsymbol{\theta}_G$ as critic we use conditional log-likelihood $\log Pr(q|p;\boldsymbol{\theta}_G)$ as the confidence score. For $\boldsymbol{\theta}_R$ as critic we use the dot product similarity between the encodings $E_P(p)$ and $E_Q(q)$ as the confidence score. Self-training and back-training can be combined with one or both of the these consistency checks. We set filter thresholds to accept 75\% of synthetic data (refer to \cref{sec:train-details} for exact threshold values).

A popular data filtering technique in data augmentation is cycle consistency \cite{alberti-etal-2019-synthetic} which is enforced by further generating noisy input from noisy output, and matching noisy input similarity with source input. We leave its exploration as future work.

\section{Domain Adaptation Evaluation}\label{experimental-setup}
As described in \Cref{sec:background}, our source domain is NaturalQuestions and the target domains are MLQuestions and PubMedQA.
We evaluate if domain adaptation helps to improve the performance compared to no adaptation.
We empirically investigate qualitative differences between self-training and back-training to validate their effectiveness.
We also investigate if consistency filters and iterative refinement result in further improvements.

\subsection{No-adaptation vs. self-training vs back-training}
In \Cref{tab:aqgm}, we compare the performance of vanilla self-training and back-training (i.e., without consistency filtering or iterative refinement) with the no-adaptation baseline (i.e. model trained on source domain and directly tested on target domain). 
On MLQuestions, self-training achieves an absolute gain of around 0.6 BLEU-4 points for QG and 7.13 R@20 points for IR.
Whereas back-training vastly outperforms self-training, with improvements of 9.4 BLEU-4 points on QG and 19.6 R@20 points on IR over the no-adaptation baseline. The improvements are even bigger on PubMedQA whereas self-training shows no improvement at all.

\begin{figure}
\includegraphics[trim=3.3cm 18.75cm 5cm 4.2cm, clip, scale=0.825]{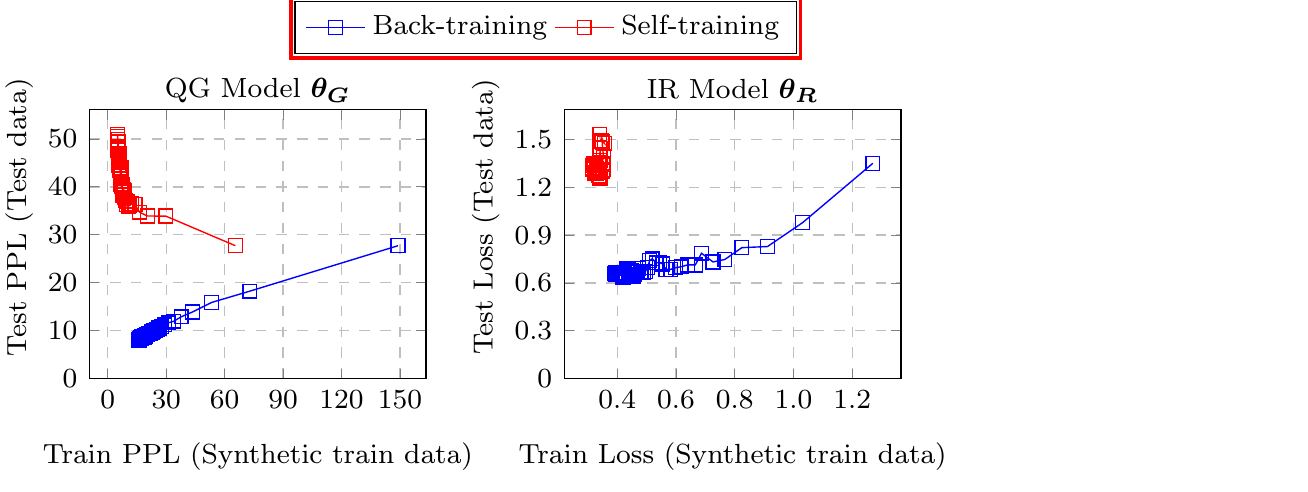}
\caption{Evolution of QG model perplexity (PPL) and IR model loss for Self-training vs Back-training as training proceeds on MLQuestions. \textit{Trajectories run from right to left as training loss decreases with time}. Rightmost points are plotted after first mini-batch training, and subsequent points are plotted after each mini-batch training.}
\label{pac-bound}
\end{figure}

\subsection{Why does back-training work?}\label{stvsct}
\Cref{pac-bound} shows the QG model perplexity and IR model loss on synthetic training data and test data as the training (domain adaptation) proceeds on     MLQuestions.
The plots reveal three interesting observations: (1) for back-training, the train and test loss (and hence likelihood) are correlated and hence the data generated by back-training matches the target distribution more closely than self-training; 
(2) self-training achieves lower training error but higher test error compared to back-training, indicating overfitting; (3) extrapolating back-training curve suggests that scaling additional unlabeled data will likely improve the model.

\begin{figure}
\includegraphics[trim=3.3cm 19.25cm 5cm 4.2cm, clip, scale=0.9]{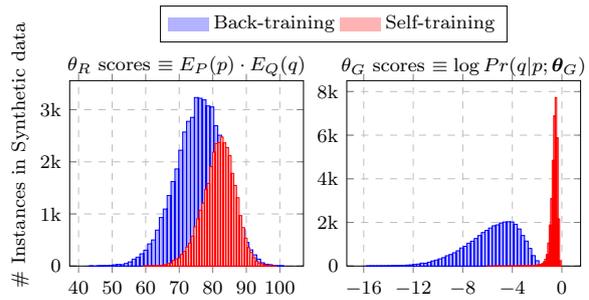}
\caption{DPR embedding similarity scores and QG Log-likelihood scores distribution on MLQuestions synthetic data computed using $\boldsymbol{\theta}_R$ and $\boldsymbol{\theta}_G$ respectively.}
\label{fig:score-distribution}
\end{figure}

Figure \ref{fig:score-distribution} plots the distribution for self-training (computing likelihood scores of model's own predictions) and back-training (computing likelihood scores of different model's predictions) for QG and IR tasks on MLQuestions.
The figures reveal that self-training curve has \textit{high mean} and \textit{low variance}, indicating less diverse training data.
On the other hand, back-training curve has \textit{low mean} and \textit{high variance} indicating diverse training data.

\subsection{Are consistency filters useful?}
Table \ref{tab:consistency-performance} reveals that although our consistency filters outperform base models on MLQuestions, the improvements are not very significant. Our hypothesis is that quality of synthetic data is already high (as backed up by \Cref{stvsct} findings), which limits the performance gain. However, the filters reduce synthetic training data by 25\%, which leads to faster model training without any drop in performance. Additionally, self-consistency improves self-training in many problems \cite{zhu2005semi, sachan2018self}. We believe our cross-consistency filter could also be explored on similar problems.

\begin{table}[]
    \footnotesize
    \centering
    \begin{tabular}{lrrrr}
    \toprule
         & \multicolumn{2}{c}{\textbf{QG}} & \multicolumn{2}{c}{\textbf{IR}} \\
        \textbf{Consistency} & BLEU4 & ROUGE & R@20 & R@100 \\
        \midrule
        \multicolumn{5}{c}{\textit{Self-Training}} \\
        \midrule
        None & 9.43 & 32.18 & 65.26 & 83.06 \\
        Self & \textbf{9.85} & \textbf{32.34} & 64.75 & \textbf{83.23} \\
        Cross & 8.92 & 31.97 & \textbf{65.46} & 83.00 \\
        \midrule
        \multicolumn{5}{c}{\textit{Back-Training}} \\
        \midrule
        None & 18.48 & 43.97 & 77.73 & 91.73 \\
        Self & 18.62 & \textbf{44.19} & 77.40 & 91.66 \\
        Cross & 18.67 & 43.22 & \textbf{78.86} & \textbf{92.13} \\
        \bottomrule
    \end{tabular}
    \caption{Effect of using consistency filters on Self-Training and Back-Training for MLQuestions. }
    \label{tab:consistency-performance}
\end{table}

\subsection{Is iterative refinement useful?}
Further performance improvement of up to 1.53 BLEU-4 points and 2.07 R@20 points can be observed in back-training (Table \ref{tab:iterative}) via the iterative procedure described in \Cref{algo}. 
On the other hand, self-training does not show any improvements for QG and marginal improvements for IR. 

\begin{table}[]
    \footnotesize
    \centering
    \vspace{1em}
    \begin{tabular}{lllll}
    \toprule
         & \multicolumn{2}{c}{\textbf{QG}} & \multicolumn{2}{c}{\textbf{IR}} \\
        \textbf{Iteration} & BLEU4 & ROUGE & R@20 & R@100 \\
        \midrule
        \multicolumn{5}{c}{\textit{Self-Training}} \\
        \midrule
        $T=1$ & 9.43 & 32.18 & 65.26 & 83.06 \\
        $T=2$ & \textcolor{red}{9.28$\downarrow$} & \textcolor{red}{32.09$\downarrow$} & \textcolor{blue}{65.60$\uparrow$} & \textcolor{blue}{83.78$\uparrow$} \\
        \hline
        Net Gain & 0 & 0 & 0.34 & 0.72 \\
        \midrule
        \multicolumn{5}{c}{\textit{Back-Training}} \\
        \midrule
        $T=1$ & 18.48 & 43.97 & 77.73 & 91.73 \\
        $T=2$ & \textcolor{blue}{20.01$\uparrow$} & \textcolor{blue}{46.02$\uparrow$} & \textcolor{blue}{79.80$\uparrow$} & \textcolor{blue}{93.26$\uparrow$} \\
        \hline
        Net Gain & 1.53 & 2.05 & 2.07 & 1.53 \\
        \bottomrule
    \end{tabular}
    \caption{Evolution of model performance on MLQuestions with increasing iterations: \textcolor{blue}{Blue} numbers denote increases in performance, while \textcolor{red}{Red} numbers denote decrease in performance.}
    \label{tab:iterative}
\end{table}

\subsection{Human Evaluation Results}
We also report human evaluation of QG by sampling 50 generated questions from MLQuestions test set and asking three domain experts to rate a question as good or bad based on four attributes: \textit{Naturalness}, i.e., fluency and grammatical correctness; \textit{Coverage}, i.e., whether question covers the whole passage or only part of the passage; \textit{Factual Correctness} in ML domain;  \textit{Answerability}, i.e., if the question can be answered using the passage. 
From the results in Table \ref{tab:human_evalutation}, we observe that the back-training model is superior on all four criteria. 
However, all models perform similarly on \textit{naturalness}.

In Table \ref{tab:running_examples} we present some generated questions of various models on MLQuestions and PubMedQA dataset. Subjectively, we find that no-adaptation and self-training models fail to understand domain knowledge, generate generic questions and miss important words present in gold question. Whereas back-training generated question matches more closely to gold question.

\begin{table}[htp]
    \centering
    \vspace{1em}
    \begin{tabular}{ccccc}
    \toprule
        \textbf{Model} & \textbf{N} & \textbf{C} & \textbf{FC} & \textbf{A} \\
        \midrule
        No-adaptation & 0.64 & 0.30 & 0.58 & 0.68 \\
        Self-Training & 0.63 & 0.32 & 0.58 & 0.70 \\
        Back-Training & \textbf{0.66} & \textbf{0.41} & \textbf{0.64} & \textbf{0.88} \\
        \bottomrule
    \end{tabular}
    \caption{Human evaluations scores between 0-1 on 50 model generated questions for four criteria: Naturalness (N), Coverage (C), Factual Correctness (FC), and Answerability (A).}
    \label{tab:human_evalutation}
    \vspace{1em}
\end{table}

\begin{table*}[]
    \centering
    \small
    \begin{tabular}{ll}
    \toprule
        \textbf{Passage} & \textbf{Questions} \\
        \midrule 
     \textit{If the line is a good fit for the data then } & \textbf{No-adaptation:} What is the meaning of random plot in statistics? \\
        \textit{the residual plot will be random. However,} & \textbf{ST:} What is the meaning of random plot in statistics? \\
        \textit{if the line is a bad fit for the data then} & \textbf{BT:} How do you know if a residual plot is random? \\
        \textit{the plot of residuals will be random.} & \textbf{Reference:} How do you know if a residual plot is good?\\
        \midrule
        \textit{Financial incentives for smoking cessation} & \textbf{No-adaptation:} When do we stop smoking in pregnancy? \\
         \textit{in pregnancy are highly cost-effective,} & \textbf{ST:} When do you stop smoking in pregnancy? \\
        \textit{with an incremental cost per quality} & \textbf{BT:} Is there a financial incentive for smoking cessation in pregrancy?\\
        \textit{adjusted life years of £482, which is well} & \textbf{Reference:} Are financial incentives cost-effective to support smoking and\\
        \textit{below recommended decision thresholds.} & cessation during pregnancy?\\
        \bottomrule
    \end{tabular}
    \caption{Examples of generated questions from MLQuestions (first row) and PubMedQA (second row). ST and BT refer to Self-training and Back-training models respectively.}
    \label{tab:running_examples}
\end{table*}

\subsection{Analysis of Question Types}

\begin{figure}[ht]
    \centering
    \includegraphics[trim=3cm 17.5cm 5cm 4.1cm, clip,scale=0.9]{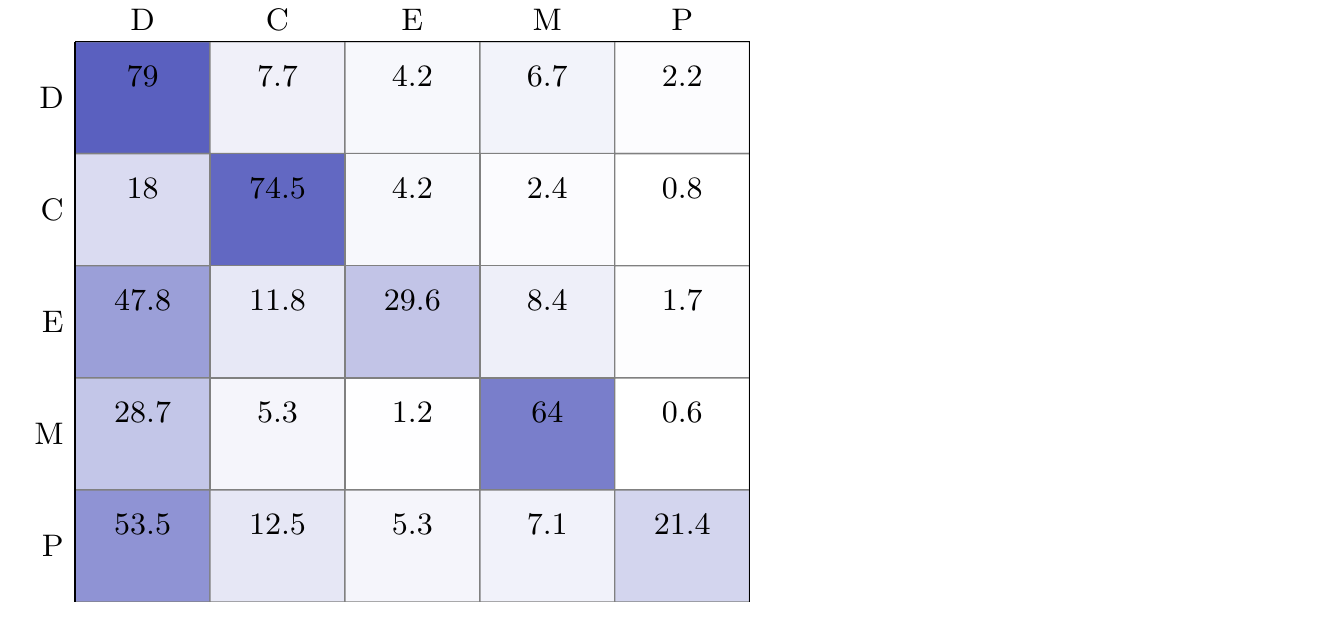}
    \caption{Confusion matrix of actual (row) vs model generated question (column) classes for 100 questions sampled from MLQuestions test set. Classes are abbreviated as Description (D) Comparison (C), Explanation (E), Method (M), and Preference (P). Values are in \% where each row sums to 100\%.}
    \label{fig:confusion-matrix-taxonomy}
    \vspace{1em}
\end{figure}

We analyze how well our QG model can generate different kinds of questions according to the taxonomy described in Table \ref{tab:taxonomy-table}. 
In \Cref{fig:confusion-matrix-taxonomy} we plot the confusion matrix between the actual question class and generated question class for our back-training model. To do this, 100 actual questions and corresponding generated questions are sampled from the MLQuestions test set and annotated by a domain expert. We find that the model generates few \textit{Explanation} questions and even fewer \textit{Preference} questions while over-generating \textit{Description} questions.
\textit{Comparison} and \textit{Method} questions show good F1-score overall, hence these classes benefit the most from domain adaptation.

\section{Related Work}

\paragraph{Question Generation} methods have focused on training neural Seq2Seq models \cite{du2017learning, mishra2020towards,zhao2018paragraph,chan2019bert,klein2019learning} on supervised QA datasets such as SQuAD \cite{rajpurkar2016squad}. Many recent works such as \cite{tang2017question, wang2017joint} recognize the duality between QG and QA and propose joint training for the two. 
\newcite{duan2017question} generate QA pairs from YahooAnswers, and improve QA by adding a question-consistency loss in addition to QA loss.
Our work instead establishes strong duality between QG and IR task.
Ours is also the first work towards unsupervised domain adaptation for QG to the best of our knowledge.

\paragraph{Passage Retrieval} has previously been performed using classical Lucene-BM25 systems \cite{robertson2009probabilistic} based on sparse vector representations of question and passage, and matching keywords efficiently using TF-IDF. 
Recently, \citet{karpukhin2020dense} show that fine-tuning dense representations of questions and passages on BERT outperforms classical methods by a strong margin. 
We adopt the same model for domain adaptation of IR.
Concurrent to our work, \citet{reddy2020end} also perform domain adaptation for IR.
Our focus has been on systematically approaching UDA problem for both QG and IR.

\paragraph{Data Augmentation} methods like self-training have been applied in numerous NLP problems such as question answering \cite{chung2018supervised}, machine translation \cite{ueffing2006self}, and sentiment analysis \cite{he2011self}. 
\newcite{sachan2018self} apply self-training to generate synthetic data for question generation and question answering (QA) in the same domain, and filter data using QA model confidence on answer generated by question.

Back-translation's idea of aligning real outputs with noisy inputs is shared with back-training and has been successful in improving Unsupervised NMT \cite{artetxe2018unsupervised, edunov2018understanding}. \newcite{zhang2018style} use back-translation to generate synthetic data for the task of automatic style transfer. Back-training also shares similarities with co-training \cite{blum1998combining, wan2009co} and tri-training \cite{li-etal-2014-ambiguity,weiss-etal-2015-structured} where multiple models of \textit{same} task generate synthetic data for each other.

\section{Conclusion and Future Work}
We introduce back-training as an unsupervised domain adaptation method focusing on Question Generation and Passage Retrieval. Our algorithm generates synthetic data pairing high-quality outputs with noisy inputs in contrast to self-training producing noisy outputs aligned with quality inputs. We find that back-training outperforms self-training by a large margin on our newly released dataset MLQuestions and PubMedQA. 

One area of future research will be exploring back-training for other paired tasks like visual question generation \cite{mostafazadeh2016generating} and image retrieval \cite{datta2008image}, and style transfer \cite{gatys2015neural} from source to target domain and vice-versa.
The theoretical foundations for the superior performance of back-training have to be explored further.

\newpage 

\section*{Acknowledgments}
We thank the members of SR's research group for their constant feedback during the course of work. 
We thank Ekaterina Kochmar, Ariella Smofsky and Shayan from Korbit ML team for their helpful comments. 
SR is supported by the Facebook CIFAR AI Chair and the NSERC Discovery Grant. 
DK is supported by the MITACS fellowship.
We would like to thank SerpAPI for providing search credits for data collection in this work.

\bibliography{custom}
\bibliographystyle{acl_natbib}

\newpage
\newpage

\appendix
\section{Appendix}

\subsection{Model Training Details} \label{sec:train-details}
All experiments are run with same training configuration. Mean scores across 5 individual runs are provided on the test set. We describe the full model training details below for reproducibility.\\\\
\textbf{BART Question Generation Transformer}\\
We train BART-Base\footnote{We use huggingface BART implementation \url{https://huggingface.co/transformers/model_doc/bart.html}} with batch size 32 and learning rate of 1e-5. For all experiments we train the model for 5 epochs, though the model converges in 2-3 epochs. For optimization we use Adam \cite{kingma2015adam} with $\beta_1=0.9, \beta_2=0.999, \epsilon=1e-8$. The question and passage length is padded to 150 and 512 tokens respectively. For decoding we use top-k sampling \cite{fan2018hierarchical} with $k=50$. The model is trained with standard cross-entropy objective. \\\\
\textbf{Dense Passage Retriever (DPR)}\\
We use publicly available implementation of DPR model\footnote{\url{https://github.com/facebookresearch/DPR}} to train our IR system. We also use pre-trained NQ DPR checkpoint provided by authors \footnote{\url{https://github.com/facebookresearch/DPR/blob/master/dpr/data/download_data.py}} as the model trained on source domain of NaturalQuestions dataset. The model is trained for 5 epochs with batch size of 32 for all experiments with default hyperparameter settings in \citet{karpukhin2020dense}. \citet{karpukhin2020dense} also construct negative examples for each (passage, question) pair where the model maximizes question similarity with gold passage and minimizes similarity with negative passages simultaneously. We construct negative passages similar to \citet{karpukhin2020dense} as the top-k passages returned by BM25 which match most question tokens but don't contain the answer. We set $k=7$ for our experiments. For iterative refinement models, we always use same negative passages as the model obtained after 1st iteration ($T=1$). This is because after each iteration model is being \textit{fine-tuned} starting from previous model and not \textit{re-trained} on pseudo-data. We obtain better performance gains on dev set following this setting.\\\\
\textbf{Consistency Filters}\\
\Cref{tab:consistency-thresholds} enlists threshold values for different consistency filters. Values are arrived at by plotting confidence scores distribution of synthetic data, and setting threshold to accept 75\% of the data (i.e. third quartile Q3). As explained in \cref{sec:filters}, for $\boldsymbol{\theta}_G$ as critic we use conditional log-likelihood $\log Pr(q|p;\boldsymbol{\theta}_G)$ as our confidence scores. For $\boldsymbol{\theta}_R$ as critic we use DPR similarity score $E_P(p)E_Q(q)$ as our confidence scores. 

\begin{table}[]
    \centering
    \begin{tabular}{ccc}
    \toprule
         & \multicolumn{2}{c}{\textbf{Critic}} \\
        \textbf{Consistency} & $\boldsymbol{\theta}_G$ & $\boldsymbol{\theta}_R$ \\
        \midrule
        Self consistency & -1.19 & 78.24 \\
        Cross consistency & -5.95 & 71.65 \\
        \bottomrule
    \end{tabular}
    \caption{Threshold values for different consistency filters. Values are chosen as the third quartile (Q3) of score distribution of synthetic data, accepting 75\% of synthetic data for model training.}
    \label{tab:consistency-thresholds}
\end{table}

\subsection{NaturalQuestions Dataset Pre-processing}\label{sec:nq-preprocess}
We use Google NaturalQuestions dataset as our \textit{source} domain corpus. We pre-process publicly available train and dev corpora in a similar manner to \cite{mishra2020towards} by selecting all questions starting from the \textit{long-answer} tag and filtering out cases where the long-answer doesn't start with the HTML <p> tag. We obtain 108,501 examples which we split into a 90/10 ratio for training/dev sets. The NQ dev set of 2,136 examples is used as our test data (as the test set is hidden). 

\begin{figure}
    \centering
    \includegraphics[scale=0.8]{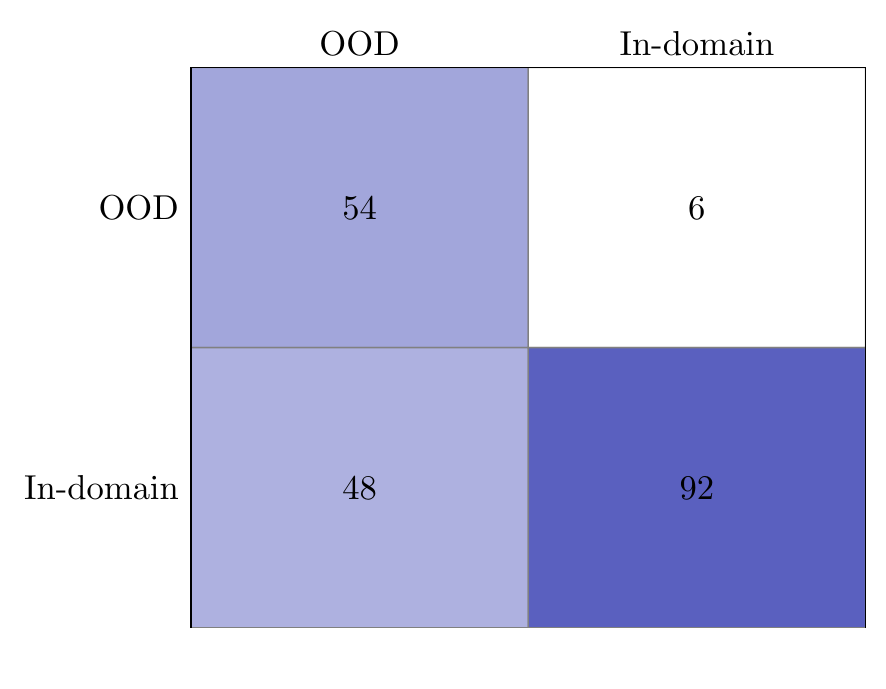}
    \caption{Test set Confusion matrix of Out-of-domain (OOD) and In-domain classes for classifier probability threshold of 0.8.}
    \label{fig:ood-filtering-confusion-matrix}
\end{figure}
\begin{figure}
    \centering
    \includegraphics[scale=0.5]{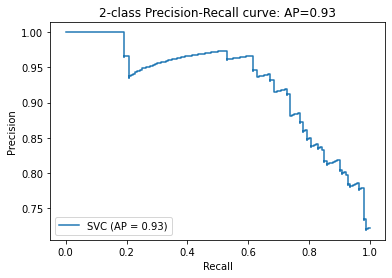}
    \caption{Precision-Recall curve for Test set of 150 questions. AP denotes average precision.}
    \label{fig:ood-filtering-precision-recall-curve}
\end{figure}

\subsection{MLQuestions: Filtering undesirable data}\label{ood_clf}
This section describes filtering out-of-domain questions (OOD) from collected 104K questions from Google described in \cref{sec:mlquestions}. Many ML terms are \textit{homonyms} \cite{menner1936conflict}: they have a different meaning in another context - (e.g.\@ ``Ensemble'', ``Eager Learning'', ``Transformers''). This means the collected data contains OOD questions. Upon analyzing $100$ random questions drawn from 104K questions, we find 27 of them are OOD. 

To filter such undesirable data, we randomly sample 1000 questions and recruit 3 domain experts to label them as In-domain or OOD. 200 questions were labeled by all 3 to determine inter-annotator agreement. We record a Cohen's Kappa agreement score \cite{mchugh2012interrater} of 0.84. The 1000 annotated questions are split into sizes 800, 50, 150 for train, dev, and test sets respectively. Based on this labeled data, we train a classifier on top of question features to classify remaining questions as \textit{useful} or \textit{OOD}. For extracting features from questions, we utilize DistillBERT model \cite{sanh2019distilbert} trained on SNLI+MultiNLI \cite{bowman2015large, williams2018broad} and then fine-tuned on the STS benchmark\cite{cer2017semeval} train set\footnote{We use off-the-shelf implementation \url{https://github.com/UKPLab/sentence-transformers} to extract sentence features from pretrained model}. This gives us feature vector of size $768$ which is used to train SVM classifier\footnote{\url{https://scikit-learn.org/stable/modules/svm.html}} with L2 penalty of $0.1$. We carefully set the acceptance threshold relatively high to $0.8$, to ensure high precision, thus accepting very few OOD questions.

Figure \ref{fig:ood-filtering-confusion-matrix} shows confusion matrix on test set with $\alpha$ set as $0.8$. The classifier obtains high precision and average recall of $94.6\%$ and $66\%$ respectively. High precision is empirically verified by annotating 100 random accepted questions, out of which 92 are found to be in-domain. The remaining 8\% of the data can be treated as noise for model training. Figure \ref{fig:ood-filtering-precision-recall-curve} plots the precision-recall trade-off by varying the acceptance threshold $\alpha$.

\subsection{Taxonomy of MLQuestions Dataset}\label{sec:taxonomy}
In Table \ref{tab:taxonomy-table}, we show the distribution of various types of questions in MLQuestions and NaturalQuestions dataset. We split the questions into 5 categories based on Nielsen's Educational Taxonomy \cite{nielsen2008taxonomy}: \textit{descriptive} questions, which ask for definitions or examples; \textit{method} questions which ask for computations or procedures; \textit{explanation} questions, which ask for justifications; \textit{comparison} questions, which ask to compare two or more concepts; and \textit{preference} questions, which are answered by a selection from a set of options.
Refer to \citet{nielsen2008question} for detailed understanding of the taxonomy.

\section{Reproducibility Checklist}
\subsection{For all reported experimental results}
\begin{itemize}
    \item \textit{A clear description of the mathematical setting, algorithm, and/or model:} This is provided in \Cref{sec:background} and \Cref{sec:train-details} of the main paper.
    \item \textit{Submission of a zip file containing source code, with specification of all dependencies, including external libraries, or a link to such resources (while still anonymized):} We provide the source code zipped repository \textit{MLQuestions}. The README file contains all instructions needed to replicate experiments. The file \textit{requirements.txt} specifies required python dependencies.
    \item \textit{Description of computing infrastructure used:} We perform our experiments on a machine with specifications: 2 CPUs, 2 RTX8000 GPUs, 18GB RAM.
    \item \textit{The average runtime for each model or algorithm (e.g., training, inference, etc.), or estimated energy cost:} \Cref{tab:runtime} lists average runtime for each step of vanilla Self-training and back-training algorithms, as well as for consistency filters.
    \item \textit{Number of parameters in each model:} For question generation, the BART base model contains total 139M parameters. For passage retrieval, the DPR model contains total 220M parameters.  
    \item \textit{Corresponding validation performance for each reported test result:} Tables \ref{tab:aqgm-dev}, \ref{tab:consistency-performance-dev}, \ref{tab:dev-iterative} report the validation set performance for each reported test result in the main paper.
    \item \textit{Explanation of evaluation metrics used, with links to code:} Refer to \Cref{sec:eval-metrics} of main paper for explanation of evaluation metrics. For evaluating QG model, we use the Maluuba NLG-Eval github library to compute BLEU, METEOR, ROUGE scores. The repository can be found at \\ \url{https://github.com/Maluuba/nlg-eval}. For IR, we implement the top-K retrieval accuracy which can be found in the file location \url{file://MLQuestions/IR/eval_retriever.py} of our submitted source code.
\end{itemize}

\begin{table}[]
    \centering
    \small
    \begin{tabular}{lcr}
    \toprule
        \textbf{Task} & \textbf{Data Size} & \textbf{Runtime} \\
        \midrule
        $\boldsymbol{\theta_G}$ generates synthetic data & 50K & 174 \\
        $\boldsymbol{\theta_R}$ generates synthetic data & 35K & 110 \\
        $\boldsymbol{\theta_G}$ filters low-quality data & 35K & 31 \\
        $\boldsymbol{\theta_G}$ filters low-quality data & 50K & 45 \\
        $\boldsymbol{\theta_G}$ filters low-quality data & 35K & 35 \\
        $\boldsymbol{\theta_G}$ filters low-quality data & 50K & 48 \\
        $\boldsymbol{\theta_G}$ Self-training & 50K & 373 \\
        $\boldsymbol{\theta_G}$ Back-training & 35K & 263 \\
        $\boldsymbol{\theta_R}$ Self-training & 35K & 547 \\
        $\boldsymbol{\theta_R}$ Back-training & 50K & 762 \\
        \bottomrule
    \end{tabular}
    \caption{Runtime (in minutes) for each step in domain adaptation models for MLQuestions dataset. Since there are 35K unaligned questions and 50K unaligned passages, a step has different execution times depending on type of training (self/back) or consistency filter (self/cross).}
    \label{tab:runtime}
\end{table}

\begin{table*}[ht]
\setlength{\tabcolsep}{5.25pt}
\footnotesize
    \centering
    \begin{tabular}{c|c|cccccc|cccc}
    \toprule
        \multirow{2}{*}{\textbf{Dataset}} & \multirow{2}{*}{\textbf{Model}} & \multicolumn{6}{c|}{Question Generation} & \multicolumn{4}{c}{Passage Retrieval} \\
        \cline{3-12}
         & & B1 & B2 & B3 & B4 & M & R & R@1 & R@20 & R@40 & R@100 \\
        \hline
       \multirow{3}{*}{\textit{MLQuestions}} & No-adaptation & 30.64 & 19.70 & 12.82 & 8.80 & 23.23 & 31.33 & 13.00 & 54.86 & 64.6 & 73.93 \\
        & Self-Training & 31.01 & 20.36 & 13.50 & 9.37 & 23.67 & 31.75 & 14.13 & 62.20 & 70.80 & 80.66 \\
        & \textbf{Back-Training} & \textbf{41.42} & \textbf{30.72} & \textbf{22.50} & \textbf{17.29} & \textbf{23.38} & \textbf{41.58} & \textbf{21.86} & \textbf{77.40} & \textbf{84.66} & \textbf{90.26} \\
        \hline
        \multirow{3}{*}{\textit{PubMedQA}} & No-adaptation & 14.23 & 7.02 & 3.65 & 1.81 & 9.12 & 15.96 & 32.66 & 57.20 & 61.8 & 72.68 \\
        & Self-Training & 14.04 & 6.98 & 3.09 & 1.54 & 8.67 & 15.30 & 33.0 & 57.48 & 64.2 & 73.44 \\
        & \textbf{Back-Training} & \textbf{27.17} & \textbf{17.92} & \textbf{12.34} & \textbf{8.76} & \textbf{17.66} & \textbf{25.89} & \textbf{56.60} & \textbf{81.0} & \textbf{83.20} & \textbf{87.68}
        \\
        \bottomrule
        \hline
    \end{tabular}
    \caption{Validation set results of unsupervised domain adaptation. \textit{No-adaptation} denotes the model trained on NaturalQuestions and evaluated directly on MLQuestions/PubMedQA dev sets without any domain adaptation.}
    \label{tab:aqgm-dev}
\end{table*}

\begin{table}
    \footnotesize
    \centering
    \begin{tabular}{lrrrr}
    \toprule
         & \multicolumn{2}{c}{\textbf{QG}} & \multicolumn{2}{c}{\textbf{IR}} \\
        \textbf{Consistency} & BLEU4 & ROUGE & R@20 & R@100 \\
        \midrule
        \multicolumn{5}{c}{\textit{Self-Training}} \\
        \midrule
        None & 9.37 & 31.75 & 62.20 & 80.66 \\
        Self & \textbf{9.76} & 31.67 & 62.75 & 81.56 \\
        Cross & 9.02 & \textbf{32.34} & \textbf{62.96} & \textbf{82.00} \\
        \midrule
        \multicolumn{5}{c}{\textit{Back-Training}} \\
        \midrule
        None & 17.29 & 41.58 & 77.40 & 90.26 \\
        Self & 17.91 & \textbf{43.27} & 76.86 & 91.06 \\
        Cross & \textbf{18.09} & 41.84 & \textbf{78.26} & \textbf{91.33} \\
        \bottomrule
    \end{tabular}
    \caption{Effect of using consistency filters on Self-Training and Back-Training for MLQuestions validation set. }
    \label{tab:consistency-performance-dev}
\end{table}

\begin{table}[]
    \footnotesize
    \centering
    \begin{tabular}{lllll}
    \toprule
         & \multicolumn{2}{c}{\textbf{QG}} & \multicolumn{2}{c}{\textbf{IR}} \\
        \textbf{Iteration} & BLEU4 & ROUGE & R@20 & R@100 \\
        \midrule
        \multicolumn{5}{c}{\textit{Self-Training}} \\
        \midrule
        $T=1$ & 9.37 & 31.75 & 62.20 & 80.66 \\
        $T=2$ & \textcolor{red}{9.22$\downarrow$} & \textcolor{red}{31.13$\downarrow$} & \textcolor{blue}{62.80$\uparrow$} & \textcolor{blue}{81.08$\uparrow$} \\
        \hline
        Net Gain & 0 & 0 & 0.60 & 0.42 \\
        \midrule
        \multicolumn{5}{c}{\textit{Back-Training}} \\
        \midrule
        $T=1$ & 17.29 & 41.58 & 77.40 & 90.26 \\
        $T=2$ & \textcolor{blue}{19.97$\uparrow$} & \textcolor{blue}{45.74$\uparrow$} & \textcolor{blue}{78.56$\uparrow$} & \textcolor{blue}{91.26$\uparrow$} \\
        \hline
        Net Gain & 2.68 & 4.16 & 1.16 & 1.00 \\
        \bottomrule
    \end{tabular}
    \caption{Evolution of model performance on MLQuestions validation set with increasing iterations: \textcolor{blue}{Blue} numbers denote increases in performance, while \textcolor{red}{Red} numbers denote decrease in performance.}
    \label{tab:dev-iterative}
\end{table}

\subsection{For all experiments with hyperparameter search}
\begin{itemize}
    \item \textit{The exact number of training and evaluation runs:} For all experiments we train the QG and IR for 5 epochs. We evaluate the model performance using evaluation metrics after each epoch on the validation set, and find that models converge after 2-3 epochs.
    \item \textit{Bounds for each hyperparameter:} We experimented by manually varying hyperparameters in vicinity of values mentioned in \cref{sec:train-details}. The best hyperparameters on validation set were chosen for final model training.
    \item \textit{Hyperparameter configurations for best-performing models:} We provide complete hyperparameters details for QG and IR model in \Cref{sec:train-details}.
    \item \textit{The method of choosing hyperparameter values (e.g., uniform sampling, manual tuning, etc.) and the criterion used to select among them (e.g., accuracy):} We use manual tuning method with the criterion as BLEU-4 accuracy for QG and R@40 retrieval accuracy for IR task on validation set.
    \item \textit{Summary statistics of the results (e.g., mean, variance, error bars, etc.):} Mean scores across 5 individual runs are provided for all experiments of main paper.
\end{itemize}

\subsection{For all datasets used}
\begin{itemize}
    \item \textit{Relevant details such as languages, and number of examples and label distributions:} \Cref{sec:background} provide statistics of NaturalQuestions, MLQuestions, and PubMedQA datasets. All datasets are in English language.
    \item \textit{Details of train/validation/test splits:} This is also provided in \cref{sec:background} for all three datasets.
    \item \textit{Explanation of any data that were excluded, and all pre-processing steps:} Relevant details are provided in \cref{sec:background} for all three datasets.
    \item \textit{A zip file containing data or link to a downloadable version of the data:} We provide MLQuestions dataset in the submission zip file. The NaturalQuestions and PubMedQA dataset can be downloaded from \url{https://ai.google.com/research/NaturalQuestions/download} and \url{https://github.com/pubmedqa/pubmedqa} repsectively. The datasets can be pre-processed following the procedures mentioned in \cref{sec:background}.
    \item \textit{For new data collected, a complete description of the data collection process, such as instructions to annotators and methods for quality control.:} We provide above details for our newly created dataset \textit{MLQuestions} in \cref{sec:mlquestions}.
\end{itemize}

\end{document}